%% file: main.tex
\documentclass[conference]{IEEEtran}
\IEEEoverridecommandlockouts
\usepackage{cite}
\usepackage{bm}
\usepackage{amsfonts,amssymb}
\usepackage{amsmath,amssymb,amsfonts}
\usepackage{tikz}  
\usepackage{algorithmic}
\usepackage{graphicx}
\usepackage{mathrsfs}
\usepackage{textcomp}
\usepackage{setspace}
\usepackage{tikz}  
\usepackage{tikz} 
\usepackage{mathrsfs}
\usepackage{arydshln}
\usepackage{enumitem}
\usepackage{threeparttable}
\definecolor{custompink}{HTML}{f8a4d8}
\definecolor{customgreen}{HTML}{6aa84f}
\usepackage{array} 
\usepackage{booktabs,makecell, multirow, tabularx}
\def\BibTeX{{\rm B\kern-.05em{\sc i\kern-.025em b}\kern-.08em
    T\kern-.1667em\lower.7ex\hbox{E}\kern-.125emX}}

\newcommand{\sysname}{GDDA}

\newtheorem{problem}{Problem}

\begin{document}

 \title{\sysname{}: Semantic OOD Detection on Graphs under Covariate Shift via Score-Based Diffusion Models
\thanks{* Corresponding author}
}
\author{
\IEEEauthorblockN{Zhixia He$^1$, Chen Zhao$^2$, Minglai Shao$^{1}$\textsuperscript{*}, Yujie Lin$^{1}$, Dong Li$^{3}$, Qin Tian$^{3}$}
\IEEEauthorblockA{
$^1$\textit{School of New Media and Communication, Tianjin University, Tianjin, China} \\
$^2$\textit{Department of Computer Science, Baylor University, Waco, Texas, USA} \\
$^3$\textit{College of Intelligence and Computing, Tianjin University, Tianjin, China} \\
\{2023245033, shaoml, linyujie\_22, ld2022244154, tianqin123\}@tju.edu.cn, chen\_zhao@baylor.edu}
}

\maketitle

\begin{abstract}
    \input{abstract}

\end{abstract}


\begin{figure}[t!]
\centering
\includegraphics[width=0.85\linewidth]{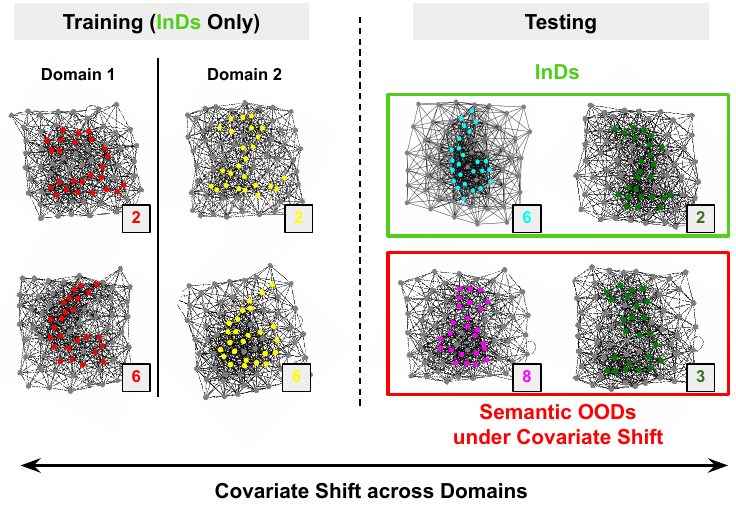}
\vspace{-3mm}
\caption{
Illustration of InD and semantic OOD graphs under covariate shift. 
In the \texttt{GOOD-CMNIST} dataset \cite{gui2022good}, digit numbers represent the semantic classes, and digit colors represent domain variations. 
The training graphs span multiple domains (red and yellow), each containing known classes (2 and 6). 
Since the testing graphs are unknown and inaccessible during training, a semantic OOD detector distinguishes in-distribution graphs (InDs) with known classes from semantic out-of-distribution graphs (OODs) with unknown classes (8 and 3), disregarding variation differences between training and testing domains.
}
\label{fig:example}
\vspace{-5mm}
\end{figure}

\section{Introduction}
\label{sec:intro}
    \input{introduction}

\section{Methodology}
\label{sec:pagestyle}

Given a set of graphs \(\mathcal G^e = \{G^e_i\}_{i=1}^{|\mathcal N_e|}\) sampled from the domain \(e\in \mathcal E\), each graph \(G\) is defined by its node feature \( X\in {\rm \mathbb R}^{\mathcal{|V|}\times d}\) and the adjacency matrix \( A \in {\rm \mathbb R}^{\mathcal{|V|}\times\mathcal{|V|}}\) as \(G=(X,A)\), where {\small\( |\mathcal V|\)} is the number of nodes and \(d\) denotes the dimension of the node features. In the graph-level OOD detection, each graph has a label \(y \in \mathcal Y\), where \(\mathcal Y\) denotes the label space. The training data and test data are drawn from different distributions, \textit{i.e.}, \(\mathbb P^{tr}( G, Y)\ne \mathbb P^{te}( G, Y)\). 

\textbf{Problem Formulation.} We assume data are sampled from multiple domains \(\mathcal E\), where the training domains denote as  \(\mathcal E_{tr}\subset \mathcal E\).  We classify  test instances with known classes as InDs and instances with the simultaneous occurrence of semantic and  covariate shift as OODs, while excluding the case where only semantic shift occurs. Our goal is to learn a semantic OOD detector that is capable of distinguishing  OODs with unknown classes from InDs in test domains, which remain invisible during the training phase.

\begin{problem}[Graph-level Semantic OOD Detection under Covariate Shift]
    Given the training domains $\mathcal{E}_{tr}\subset\mathcal{E}$, for each \(e \in \mathcal E_{tr}\), we have the corresponding data \(\mathcal D^e_{tr} = \{(G_i^e,y_i^e)\}_{i=1}^{|\mathcal N_e|}\) sampled from the distribution \(\mathbb P^{tr}(G^e,Y^e)\). Our goal is to learn a energy-based semantic detector, which can detect OODs with semantic shift while maintaining robustness under covariate shift. We train our model using the following objective: 
    \begin{equation}
    \underset{\theta \in \Theta}\min \  \underset{e \in \mathcal E_{tr}}\max\ \mathbb E_{(G^e,y^e)\sim\mathcal D^e_{tr}}\mathcal L_{cls}(\omega(f_\theta(G^e)),y^e)+\lambda \cdot \mathcal L_{energy}.
    \end{equation}   
    where {\(\omega(f_\theta(G^e))\)} is the logit output of the predictor \(f_\theta(\cdot)\). \(\mathcal L_{cls}\) denotes the cross-entropy loss of the instances in training domains, \(\mathcal L_{energy}\) represents the energy loss of InDs and auxiliary OODs, and \(\lambda\) is the hyperparameter.
\end{problem}

\textbf{Disentanglement for Covariate Shift.}
\label{phase1}
\input{phase1}

\textbf{Generalization of Auxiliary Graph Representations for Semantic OOD Detection.}
\label{phase2}
\input{phase2}

\begin{table*}[htbp]
\begin{threeparttable}
\scriptsize
\caption{OOD detection performance in terms of AUROC, AUPR, and FPR95 (mean \(\pm\) std). The best results are bolded, and \(\uparrow\) (\(\downarrow\)) indicates that the larger (smaller) values are better. InD classification accuracy (InD-Acc) is for reference only.}
\renewcommand\arraystretch{0.9}
\setlength{\tabcolsep}{4pt}
\centering
\begin{tabular}{lcccccccccccc}
  \toprule
   & \multicolumn{4}{c}{\texttt{GOOD-CMNIST}} & \multicolumn{4}{c}{\texttt{GOOD-SST2}} & \multicolumn{4}{c}{\texttt{ogbg-molbbbp}} \\ 
  \cmidrule(lr){2-5} \cmidrule(lr){6-9}\cmidrule(lr){10-13}
  &AUROC$\uparrow$ & AUPR$\uparrow$ & FPR95$\downarrow$ & InD-Acc$\uparrow$ &AUROC$\uparrow$ & AUPR$\uparrow$ & FPR95$\downarrow$&InD-Acc$\uparrow$  &AUROC$\uparrow$ & AUPR$\uparrow$ & FPR95$\downarrow$& InD-Acc$\uparrow$\\
  \midrule
  IRM \cite{arjovsky2019invariant}& 67.95\tiny$\pm$1.03\normalsize & 73.19\tiny$\pm$0.93\normalsize & 86.69\tiny$\pm$1.29\normalsize 
  & 41.48\tiny$\pm$2.71\normalsize
  &45.43\tiny$\pm$1.71\normalsize & 49.29\tiny$\pm$2.70\normalsize & 96.99\tiny$\pm$1.49\normalsize& 
  85.09\tiny$\pm$1.24\normalsize
  & 29.50\tiny$\pm$6.07\normalsize& 39.05\tiny$\pm$2.25\normalsize& 98.95\tiny$\pm$2.32\normalsize
  &
  54.84\tiny$\pm$0.90\normalsize\\
  
  ERM \cite{wu2022handling}& 55.59\tiny$\pm$1.05\normalsize & 64.54\tiny$\pm$0.97\normalsize & 95.40\tiny$\pm$0.66\normalsize & 
   40.99\tiny$\pm$2.37\normalsize
  & 46.50\tiny$\pm$1.19\normalsize & 50.19\tiny$\pm$2.77\normalsize & 96.79\tiny$\pm$0.51\normalsize& 
  77.69\tiny$\pm$0.58\normalsize
  &30.22\tiny$\pm$5.02\normalsize& 39.33\tiny$\pm$1.76\normalsize& 98.95\tiny$\pm$2.33\normalsize&
  56.60\tiny$\pm$7.91\normalsize\\
 
  DIR \cite{wu2022discovering}& 45.64\tiny$\pm$1.23\normalsize & 53.86\tiny$\pm$1.07\normalsize & 91.57\tiny$\pm$1.04\normalsize& 
  {56.44}\tiny$\pm$1.51\normalsize
  & 49.31\tiny$\pm$1.27\normalsize & 50.49\tiny$\pm$3.39\normalsize & 95.19\tiny$\pm$1.98\normalsize&
  81.28\tiny$\pm$1.24\normalsize
  & 41.54\tiny$\pm$7.87\normalsize& 49.56\tiny$\pm$6.83\normalsize& 97.87\tiny$\pm$2.23\normalsize&
  52.19\tiny$\pm$3.47\normalsize\\
  
  Mixup \cite{zhang2017mixup}& 57.65\tiny$\pm$1.21\normalsize & 63.17\tiny$\pm$1.20\normalsize & 92.28\tiny$\pm$1.14\normalsize&
  41.39\tiny$\pm$3.60\normalsize
  & 40.06\tiny$\pm$4.90\normalsize & 43.65\tiny$\pm$1.63\normalsize & 97.32\tiny$\pm$1.22\normalsize & 
  77.67\tiny$\pm$0.47\normalsize
  & 33.96\tiny$\pm$12.05\normalsize& 42.23\tiny$\pm$4.89\normalsize& 97.91\tiny$\pm$2.32\normalsize&
  53.92\tiny$\pm$0.07\normalsize\\
  
  Energy \cite{wu2023energy} & 
  75.52\tiny$\pm$2.52\normalsize &
  79.06\tiny$\pm$2.31\normalsize & 
  83.49\tiny$\pm$6.11\normalsize& 
  55.07\tiny$\pm$2.35\normalsize
  & 52.80\tiny$\pm$1.94\normalsize & 
  53.22\tiny$\pm$1.71\normalsize & 
  94.85\tiny$\pm$1.26\normalsize &
  86.05\tiny$\pm$0.51\normalsize
  & 49.32\tiny$\pm$8.52\normalsize& 51.52\tiny$\pm$6.56\normalsize& 94.79\tiny$\pm$3.89\normalsize&
  56.12\tiny$\pm$6.47\normalsize\\
  
  SCONE \cite{bai2023feed}& 39.96\tiny$\pm$0.34\normalsize & 42.82\tiny$\pm$0.40\normalsize & 99.82\tiny$\pm$0.12\normalsize& 
  45.69\tiny$\pm$1.18\normalsize
  & 47.99\tiny$\pm$3.12\normalsize & 49.65\tiny$\pm$1.53\normalsize & 95.64\tiny$\pm$1.63\normalsize& 
  70.07\tiny$\pm$1.31\normalsize& 50.01\tiny$\pm$7.13\normalsize& 52.93\tiny$\pm$6.08\normalsize& 94.07\tiny$\pm$5.50\normalsize&
  62.57\tiny$\pm$1.77\normalsize
  \\
  
  GOOD-D \cite{liu2023good}& 50.45\tiny$\pm$0.03\normalsize & 50.54\tiny$\pm$0.02\normalsize & 94.95\tiny$\pm$0.92\normalsize& 
  --\tiny\normalsize
  & 49.53\tiny$\pm$0.02\normalsize & 49.72\tiny$\pm$0.10\normalsize & 94.92\tiny$\pm$0.03\normalsize&
  --\tiny\normalsize& 28.07\tiny$\pm$0.03\normalsize& 35.73\tiny$\pm$0.01\normalsize& 95.31\tiny$\pm$0.32\normalsize&
  --\tiny\normalsize\\
  \midrule
  \sysname{} (Ours) & \textbf{77.96}\tiny$\pm$2.51\normalsize & \textbf{80.51}\tiny$\pm$2.41\normalsize & \textbf{78.89}\tiny$\pm$6.94\normalsize & 
  {51.31}\tiny$\pm$1.77\normalsize
  &\textbf{54.70}\tiny$\pm$1.16\normalsize & \textbf{54.85}\tiny$\pm$2.38\normalsize & \textbf{91.78}\tiny$\pm$1.31\normalsize&
  89.59\tiny$\pm$0.31\normalsize& \textbf{56.01}\tiny$\pm$5.96\normalsize& \textbf{51.72}\tiny$\pm$4.85\normalsize& \textbf{88.01}\tiny$\pm$11.85\normalsize&
  54.16\tiny$\pm$5.27\normalsize\\
  \bottomrule
\end{tabular}
\label{tab2}
\begin{tablenotes}
\tiny
\item The symbol ``--`` indicates that the method cannot compute InD accuracy.
\end{tablenotes}
\end{threeparttable}
 \vspace{-0.5cm}
 
\end{table*}

\begin{table}[!]
\caption{Statistics of datasets}
\vspace{-3mm}
\scriptsize
    \centering
    \begin{tabular}{cccc}
    \toprule
         & \# graphs & \# domains & \# classes\\
    \midrule
        \texttt{GOOD-CMNIST} \cite{gui2022good} & 70,000 &7&10\\
         \texttt{GOOD-SST2} \cite{gui2022good} & 70,042&55&2 \\
         \texttt{ogbg-molbbbp} \cite{hu2020open} &1,777&3&2\\
    \bottomrule
    \end{tabular}
    \label{tab1}
    \vspace{-0.4cm}
\end{table}
\section{Experiment}
\label{sec:typestyle}
\input{experiments}

\newpage
\vfill\pagebreak


\end{document}

%% file: abstract.tex
Out-of-distribution (OOD) detection poses a significant challenge for Graph Neural Networks (GNNs), particularly in open-world scenarios with varying distribution shifts.
Most existing OOD detection methods on graphs primarily focus on identifying instances in test data domains caused by either semantic shifts (changes in data classes) or covariate shifts (changes in data features), while leaving the simultaneous occurrence of both distribution shifts under-explored.
In this work, we address both types of shifts simultaneously and introduce a novel challenge for OOD detection on graphs: graph-level semantic OOD detection under covariate shift. 
In this scenario, variations between the training and test domains result from the concurrent presence of both covariate and semantic shifts, where only graphs associated with unknown classes are identified as OOD samples (OODs).
To tackle this challenge, we propose a novel two-phase framework called Graph Disentangled Diffusion Augmentation (\sysname{}). 
The first phase focuses on disentangling graph representations into domain-invariant semantic factors and domain-specific style factors. 
In the second phase, we introduce a novel distribution-shift-controlled score-based generative diffusion model that generates latent factors outside the training semantic and style spaces.
Additionally, auxiliary pseudo-in-distribution (InD) and pseudo-OOD graph representations are employed to enhance the effectiveness of the energy-based semantic OOD detector. 
Extensive empirical studies on three benchmark datasets demonstrate that our approach outperforms state-of-the-art baselines.





%% file: introduction.tex
Graphs are widely utilized in diverse real-world applications \cite{zhou2020graph} to represent complex relationships between entities. 
A critical challenge is to detect out-of-distribution graphs (OODs) that deviate from the in-distribution ones (InDs) to ensure the reliability and robustness of machine learning models. This is particularly important in high-stakes applications such as fraud detection \cite{pourhabibi2020fraud}, drug discovery \cite{gaudelet2021utilizing}, and personalized recommendations \cite{sha2021hierarchical}, where the presence of unknown graphs can significantly impact decision-making. 
Graph-level OOD detection faces unique challenges due to the presence of two types of distribution shifts: semantic shifts \cite{wang2003mining}, which occur when test graphs introduce previously unseen classes, and covariate shifts \cite{mandal2020ensuring}, where the feature distributions change across domains without altering class labels.
Effectively identifying unknown graphs under both types of shifts is essential for improving model generalization and ensuring accurate predictions in open-world environments.


To address the challenge of OOD detection under distribution shifts, three primary research directions have emerged.
Invariant learning methods \cite{chen2022ciga,sui2022causal} aim to explicitly extract invariant substructures, while data augmentation techniques \cite{zhang2017mixup,goodfellow2020generative} are used to enhance data diversity and improve model generalization by applying effective transformations to the training data. 
However, these approaches typically assume that OODs arise from only one type of distribution shift, either semantic or covariate shift, rather than accounting for the simultaneous occurrence of both.
In contrast, open-set domain generalization methods typically identify new classes in unknown domains by increasing the margin between known and unknown classes. 
Although these methods take both types of shifts into account, they often require training separate classifiers for each domain \cite{shu2021open} or exposing some OOD data during training \cite{bai2023feed}. 
Given the limitations of existing endeavors, there is a growing need for an OOD detection approach that can (1) effectively detect OODs arising from semantic shifts and (2) maintain robust and reliable performance even in the presence of covariate shifts in unknown domains.

\begin{figure*}[t!]
\centering
\includegraphics[width=\linewidth]{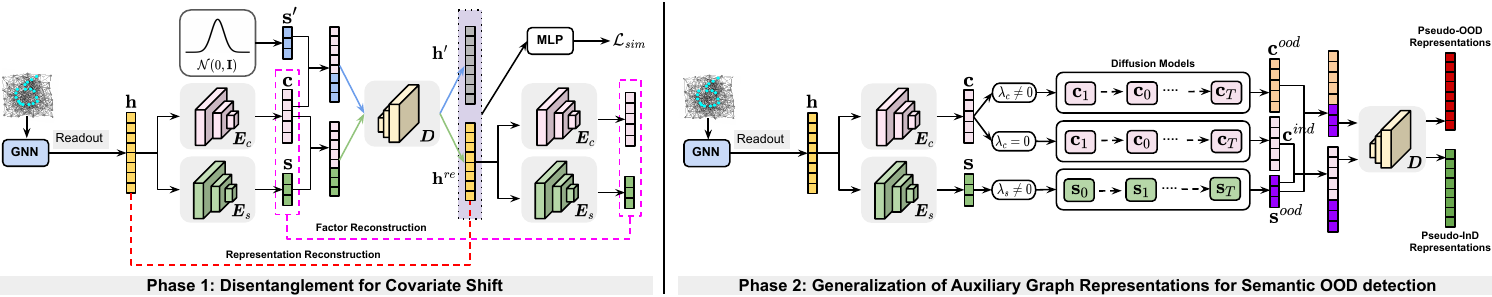}
\vspace{-7mm}
\caption{An overview of \sysname{} framework comprising two phases.  (\textbf{Left})  In the first phase, two encoders \( \bm E_c\) and \( \bm E_s\) are employed to disentangle the graph representations \(\mathbf h\) into semantic factors \(\mathbf c\) and style factors \(\mathbf s\). We sample alternative style factors \(\mathbf s'\) from Gaussian distribution, and concatenate \(\mathbf c\) with \(\mathbf s\), as well as \(\mathbf c\) with \(\mathbf s'\). These concatenated factors are then fed into the decoder \(\bm D\) for representation reconstruction. Additionally, the reconstructed representations \({\mathbf h}^{re}\) are reintroduced into encoders for factor reconstruction. 
(\textbf{Right}) In the second phase, the disentangled training factors \(\mathbf c\) and \(\mathbf s\) are incorporated into diffusion models. We apply different perturbations by setting \(\lambda_c=0\) and \(\lambda_c\neq0\) to derive the corresponding \({\mathbf c}^{ind}\) and \({\mathbf c}^{ood}\), which are then concatenated with the perturbed \({\mathbf s}^{ood}\). These concatenated factors are fed into the pre-trained decoder to generate the final pseudo representations. }
\vspace{-5mm}
\label{picture}
\end{figure*}

In this paper, we narrow the focus of OOD detection on graphs to the problem of \textit{graph-level semantic OOD detection under covariate shift}, where testing domain graphs are unknown and inaccessible during training.
An example is given in Fig. \ref{fig:example}. Given graphs from multiple training domains, each characterized by domain-specific variations (digit colors), the goal is to distinguish semantic OODs with unknown classes (8 and 3) from InDs with known classes (6 and 2) in the test domains, where all test domain graphs exhibit covariate shifts (cyan, green and purple) relative to the training domains (red and yellow).
To tackle this problem, we propose a novel framework called Graph Disentangled Diffusion Augmentation (GDDA). 
The framework consists of two phases: disentanglement for covariate shift and the generalization of auxiliary graph representations for semantic OOD detection. 
In the first phase, graph representations learned through GNNs are disentangled into domain-invariant semantic factors and domain-specific style factors. 
Additionally, to enhance the effectiveness of the energy-based semantic OOD detector, pseudo-InD and pseudo-OOD graph representations are generated using a novel distribution-shift-controlled score-based generative diffusion model, which produces latent factors beyond the training semantic and style spaces.
Our main contributions are:

\begin{itemize}[leftmargin=*]
\item We introduce a novel problem for OOD detection on graphs, referred to as graph-level semantic OOD detection under covariate shift. In this problem, all graphs in the test domains display variations different from those in the observed training domains, and the goal is to detect OODs with unknown classes from known InDs.

\item We propose a novel two-phase framework called Graph Disentangled Diffusion Augmentation (\sysname{}). The key of this framework lies in generating  latent auxiliary graph representations by introducing distribution-shift-controlled score-based generative diffusion models.


\item Empirical results on three benchmark datasets demonstrate the effectiveness of \sysname{}, showing that it significantly outperforms state-of-the-art baselines.


\end{itemize}

%% file: phase1.tex
Since each graph is defined by its structure and feature, exclusively extracting  invariant substructures may introduce unforeseen pitfalls. To obtain domain-invariant semantic and domain-specific style factors in the training semantic and style spaces, we disentangle graph-level representations in the latent space. To begin with, we adopt GNN and Readout to derive the  representations.\begin{equation}
  \small
  {\rm \mathbf{h} = {Readout}\big(\rm  GNN}(\bm G)\big) \in {\rm \mathbb{R}}^{d} \label{con:inven}
\end{equation}
\normalsize

To disentangle the representations into domain-invariant and domain-specific factors, we make use of  \(\small \bm E_c\) and \(\small \bm E_s\), which represent the semantic encoder and the style encoder.
\begin{equation}
\small
{\rm\mathbf c}=\bm E_c({\rm \mathbf h})\in {\rm \mathbb R}^{d_1}, \quad {\rm\mathbf s}=\bm  E_s({\rm \mathbf h})\in {\rm \mathbb R}^{d_2},
\end{equation}
\normalsize
where \(\mathbf c\), \(\mathbf s\) denote the semantic factors and the style factors, and \(d_1+d_2=d\).  We concatenate both components to form new factors with dimensions matching those of the original representations. These concatenated factors are then fed into the decoder  {\small\( \bm D\)}, which reorganizes the relation between the semantic and style information for representation reconstruction. 
\begin{equation}
\small \quad {\rm\mathbf h}^{re}=\bm D({\rm{{\mathbf c} \oplus \mathbf s}})\in \mathbb R^d\label{con:inventory},
\end{equation}
\normalsize

Additionally, we reintroduce \({\rm{ \mathbf h}}^{re}\) into encoders for factor reconstruction. The cooperation of  factor reconstruction and representation reconstruction enhances the disentanglement and reconstruction performance of the encoders and decoder.
\begin{equation}
\small\quad\quad \mathbf c^{ re}=\bm E_c({\rm \mathbf h}^{re})\in {\rm \mathbb R}^{d_1},\quad  \mathbf s^{ re}=\bm E_s({\rm{\mathbf h}}^{ re})\in {\rm \mathbb R}^{d_2},
\end{equation}
\normalsize
\indent The total loss is the summation of three components: (1) the reconstruction loss, (2) the similarity  loss and (3) the classification loss. First, to ensure the model can perform representation  and factor reconstruction, we compute the reconstruction loss for pairs \(\rm{\mathbf h}\) and \({\rm{\mathbf h}}^{re}\), \(\rm{\mathbf c}\) and \({\rm{\mathbf c}}^{re}\), \(\rm{\mathbf s}\) and \({\rm{\mathbf s}}^{re}\).
\begin{equation}
\small
\mathcal L_{recon}=\ell_{re}({\rm {  \mathbf h, \mathbf h}}^{re})+\ell_{re}( {\rm { \mathbf c, \mathbf c}}^{re})+\ell_{re}({\rm {\mathbf s,\mathbf s}}^{re}),\\
\end{equation}
\normalsize
where the L1Loss is utilized to compute the loss component \(\ell_{re}(\cdot,\cdot)\). Secondly, to ensure \( \rm \mathbf c\) is  the semantic factors rather than \(\rm \mathbf s\), we draw lessons from MBDG \cite{robey2021model} and take the following steps: sample  alternative style factors \(\rm{ \mathbf s'}\) from Gaussian distribution, concatenate with semantic factors \(\rm {\mathbf c}\) and feed these concatenated factors into decoder to obtain representations \(\rm {\mathbf h'}\).
\begin{equation}
\small
 { \mathbf s'}\sim\mathcal N(0,\mathbf I)\in { \mathbb R}^{d_2},\quad  {\mathbf h}' =\bm D(\mathbf c \oplus \mathbf s'),
\end{equation}
\normalsize

According to MBDG, representations \(\rm {\mathbf h}'\) and \({\rm \mathbf h}^{re}\)  should have similar softmax scores, as only the domain-specific style parts are modified. The similarity loss is calculated as follows: 
\begin{equation}
\small
\mathcal{L}_{{sim}} = d\big[s\big(\phi(\mathbf{h'})\big)\  \vert\vert \ s\big(\phi(\mathbf{h}^{ re})\big)\big],
\end{equation}
\normalsize
where {\({d}\rm[\,\cdot\,||\,\cdot\,]\)} is the Kullback-Leibler divergence and \(\rm s(\phi(\mathbf h))\) is the softmax output, with \(\phi(\mathbf h)\) serving as the intermediate logit scores.  The classification loss can be expressed as:
{\small\({\mathcal L}_{cls}= \mathbb E_{(\mathbf h',y)}\ell_{CE}\big(\phi({\rm {\mathbf h'}}),y\big)\)}, where {\small\(\ell_{CE}\)} is the cross-entropy loss. The total loss is the summation of the three parts mentioned above:
\begin{equation}
\small
 \mathcal L_{total}=\mathcal L_{recon}+\beta_1 \cdot \mathcal L_{{sim}}+\beta_2 \cdot \mathcal L_{{cls}}.
\end{equation}
\normalsize
where \(\beta_1\), \(\beta_2\) are hyperparameters that balance the three parts. After training encoders and decoder, we fix the model parameters for generating pseudo representations in phase 2.

%% file: phase2.tex
The diffusion model adds noise to data in the forward process, turns data into pure noise and then removes the noise step by step in the reverse process to recover the original input \cite{liu2023generative,ho2020denoising}.  The forward and reverse-time diffusion processes are as follows:
\begin{equation}
\small
{\rm d}\bm G_t=\mathbf f_t (\bm G_t){\rm d}t+g_t(\bm G_t){\rm d}\rm \mathbf w,
\label{equ:1}
\end{equation}
\begin{equation}
\small
{\rm d}\bm {G}_t=\Big[\mathbf f_t(\bm G_t) \ \text{--}\  g_t^2\nabla_{\bm G_t} {\rm log}p_{t}(\bm G_t)\Big]{\rm d}\bar t+g_t {\rm d}\rm \bar{\mathbf w},
\end{equation}
\normalsize
where \(g_t\) is the scalar diffusion coefficient, \(\mathbf f_t\) denotes the linear drift coefficient and {\small\(\nabla_{\bm G_t} {\rm log}p_{t}(\bm G_t)\)} is the score function.

In light of the presence of  OODs and InDs in the test domains, the semantic OOD detector should be adept at detecting semantic OODs, while ensuring robust performance under covariate shift. Our objective is to generalize the disentangled training semantic and style factors to generate auxiliary pseudo-InD and pseudo-OOD representations corresponding to InDs and OODs in test domains to enhance the effectiveness of the semantic detector. MOOD \cite{lee2023exploring} proposes a novel score-based diffusion model to generate OODs that deviate from the original distribution, where the degree of OOD-ness is controlled by a perturbation parameter. Building on this inspiration, we aim to generate latent factors outside the training semantic and style spaces with  diffusion models to narrow the gap between the training distribution and the test distribution. 

We introduce the distribution-shift-controlled score generative diffusion model to generate  the factors mentioned above beyond the training semantic and style spaces. Specifically, we introduce distribution-shift-controlled perturbation parameters \(\lambda_c\) and \(\lambda_s\) into the reverse-time diffusion process of generating OOD semantic and style factors.
Perturbing semantic factors facilitates the exploration of richer semantic information, while perturbing style factors allows for a  broader investigation of additional domain-specific information. To generate pseudo-InD representations with known classes and unknown domains, we exclusively perturb the style factors, while keeping the semantic factors intact. Conversely, to generate pseudo-OOD representations with both unknown classes and domains, we perturb both factors concurrently. The reverse-time diffusion processes for generating OOD factors are as follows:
\setlength\abovedisplayskip{8pt}
\setlength\belowdisplayskip{8pt}
\begin{equation}
\small
{\rm d}\mathbf{c}_t = \Big[ \mathbf{f}_{1,t}(\mathbf{c}_t)-(1-\lambda_c) g_{1,t}^2 \nabla_{\mathbf{c}_t}\!\log p_t(\mathbf{c}_t, \mathbf{s}_t) \Big] {\rm d}\bar{t} + g_{1,t} {\rm d}\bar{\mathbf{w}}_1,
\end{equation}
\vspace{-10pt}
\begin{equation}
\small
 {\rm d}\mathbf{s}_t = \Big[ \mathbf{f}_{2,t}(\mathbf{s}_t) - (1 - \lambda_s) g_{2,t}^2 \nabla_{\mathbf{s}_t}\!\log p_t(\mathbf{c}_t, \mathbf{s}_t) \Big] {\rm d}\bar{t} + g_{2,t} {\rm d}\bar{\mathbf{w}}_2,
\end{equation}
\normalsize
where \(\lambda_c\), \(\lambda_s\) represent separately the semantic-shift-controlled and covariate-shift-controlled perturbation parameters. We set \( \lambda_c=0\) and \( \lambda_c\neq0\) to derive factors \({\rm \mathbf c}^{ind}\) and \({\rm \mathbf c}^{ood}\), which are then concatenated with the perturbed  \({\rm \mathbf s}^{ood}\) obtained by setting \(\lambda_s\) to a non-zero constant. These concatenated factors are then fed into the pre-trained decoder to integrate and generate the pseudo-InD and pseudo-OOD augmented representations.
\begin{equation}
\small
{\rm \mathbf h}^{ind} = \bm D({\rm \mathbf {c}}^{ind}\,\oplus{\rm  \mathbf{s}}^{ood}),\qquad
{\rm \mathbf h}^{ood} = \bm D({\rm \mathbf {c}}^{ood}\,\oplus{\rm  \mathbf  {s}}^{ood}),
\end{equation}
\normalsize
where \({\rm \mathbf h}^{ind}\), \({\rm \mathbf h}^{ood}\) are the auxiliary pseudo-InD and pseudo-OOD representations. The total optimization goal is as follows:
\begin{equation}
\small
\underset{\theta \in \Theta}\min \;  \underset{e \in \mathcal E_{tr}}\max\; \mathbb E_{(G^e,y^e)\sim\mathcal D^e_{tr}}\mathcal L_{cls}\big(\omega(f_\theta(G^e)),y^e\big)+\lambda \cdot \mathcal  L_{energy}.\end{equation}\small
where 
\begin{equation} 
\begin{aligned}\small
\mathcal L_{energy}\ =\ & \mathbb E_{e\in \mathcal E_{tr}}\Big [\mathbb E_{\mathbf h^{ind}\sim \mathcal D_{ind}\cup \mathcal D_{ori}^e}\big({\rm ReLU}\big(E(\mathbf h^{ind})-m_{in}\big)\big)^2\\
+\ &\mathbb E_{\mathbf h^{ood}\sim \mathcal D_{ood}}\big({\rm ReLU}\big(E(m_{ood}-\mathbf h^{ood})\big)\big)^2\Big],
\end{aligned}
\nonumber
\end{equation}
\normalsize
and \(m_{in}\), \(m_{out}\)  are margins hyperparameters. The energy function is {\small\(E({\rm \mathbf h})=\text{--}T\cdot{\rm log} \sum_{i=1}^{N}e^{({\rm \mathbf h}/ T)}\)}, where \(T\) is the temperature parameter. \(\mathcal D_{ind}=\{\mathbf h^{ind}_i\}_{i=1}^{|\mathcal N_{ind}|}\), \(\mathcal D_{ood}=\{\mathbf h^{ood}_i\}_{i=1}^{|\mathcal N_{ood}|}\), \(\mathcal D_{ori}^e=\{\mathbf h_i^e\}_{i=1}^{|\mathcal N_{ori}|}\) are pseudo-InD, pseudo-OOD and original representations sets, where \(\mathbf h_i^e\) is sampled from domain \(e\in \mathcal E\).


%% file: experiments.tex

\textbf{Datasets.} We conduct our experiments on three datasets: \texttt{GOOD-CMNIST} \cite{gui2022good}, \texttt{GOOD-SST2} \cite{gui2022good}, and \texttt{ogbg-molbbbp} \cite{hu2020open}.
The statistics of datasets are listed in Table \ref{tab1}. 


\textbf{Baselines.}
The first group is OOD detection methods, including Mixup \cite{zhang2017mixup}, Energy \cite{wu2023energy} and GOOD-D \cite{liu2023good}. The second group is OOD generalization methods, including IRM \cite{arjovsky2019invariant}, ERM \cite{wu2022handling} and DIR \cite{wu2022discovering}. The final group is open-set domain generalization method, including  SCONE \cite{bai2023feed}.

\textbf{Evaluation metrics and settings.} We follow the common OOD detection metrics: AUROC \cite{davis2006relationship}, AUPR \cite{manning1999foundations}, FPR95 \cite{liang2017enhancing}, and InD-Acc as an auxiliary metric. For \texttt{GOOD-CMNIST} and \texttt{GOOD-SST2}, we adopt GIN-Virtual \cite{gilmer2017neural} as the GNN backbone, while for \texttt{ogbg-molbbbp}, we adopt GIN \cite{xu2018powerful}.

\textbf{Results.}
We report the results in Table \ref{tab2}, with InD-Acc as auxiliary metric for reference.  Our GDDA outperforms the leading OOD generalization baselines, with AUROC scores increased by 10.01\% on \texttt{GOOD-CMNIST}, 5.39\% on \texttt{GOOD-SST2} and 14.47\% on \texttt{ogbg-molbbbp}. Among  OOD detection algorithms, Energy  outperforms both Mixup and GOOD-D, demonstrating its robustness in handling distribution shifts. Building on the energy-based semantic OOD detector, our method achieves  AUROC score improvements of 2.44\% on \texttt{GOOD-CMNIST}, 1.90\% on \texttt{GOOD-SST2}, and 6.69\% on \texttt{ogbg-molbbbp} compared with Energy. However, as an unsupervised OOD detection method, GOOD-D is incapable of calculating the accuracy of InDs.

\begin{figure}[t!]
    \begin{minipage}[t]{0.32\linewidth}
        \centering
        \includegraphics[width=\textwidth]{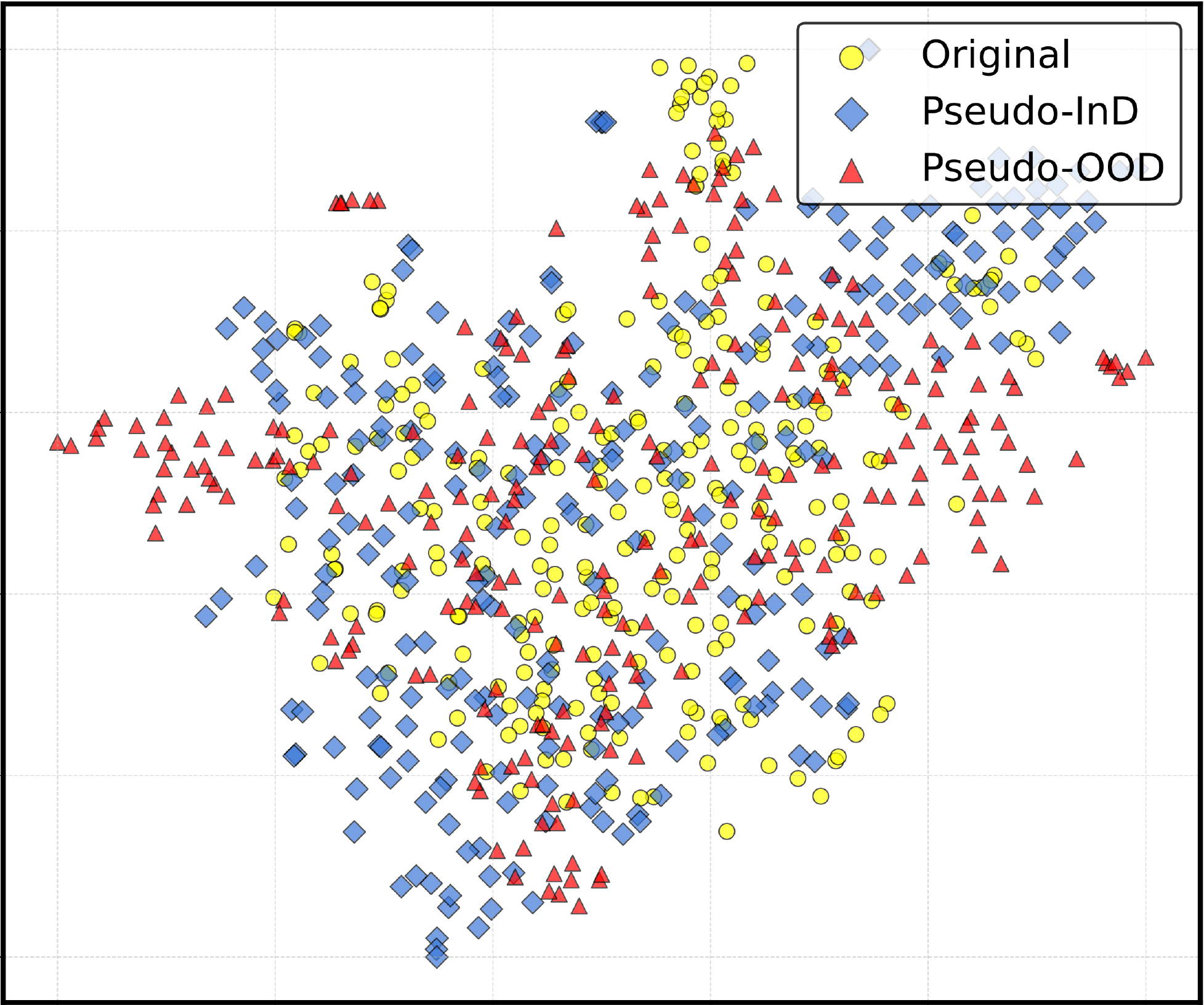}
        \centerline{(a) \footnotesize$\lambda$= 0.1}
    \end{minipage}%
    \hspace{0.0005\linewidth}
    \begin{minipage}[t]{0.32\linewidth}
        \centering
        \includegraphics[width=\textwidth]{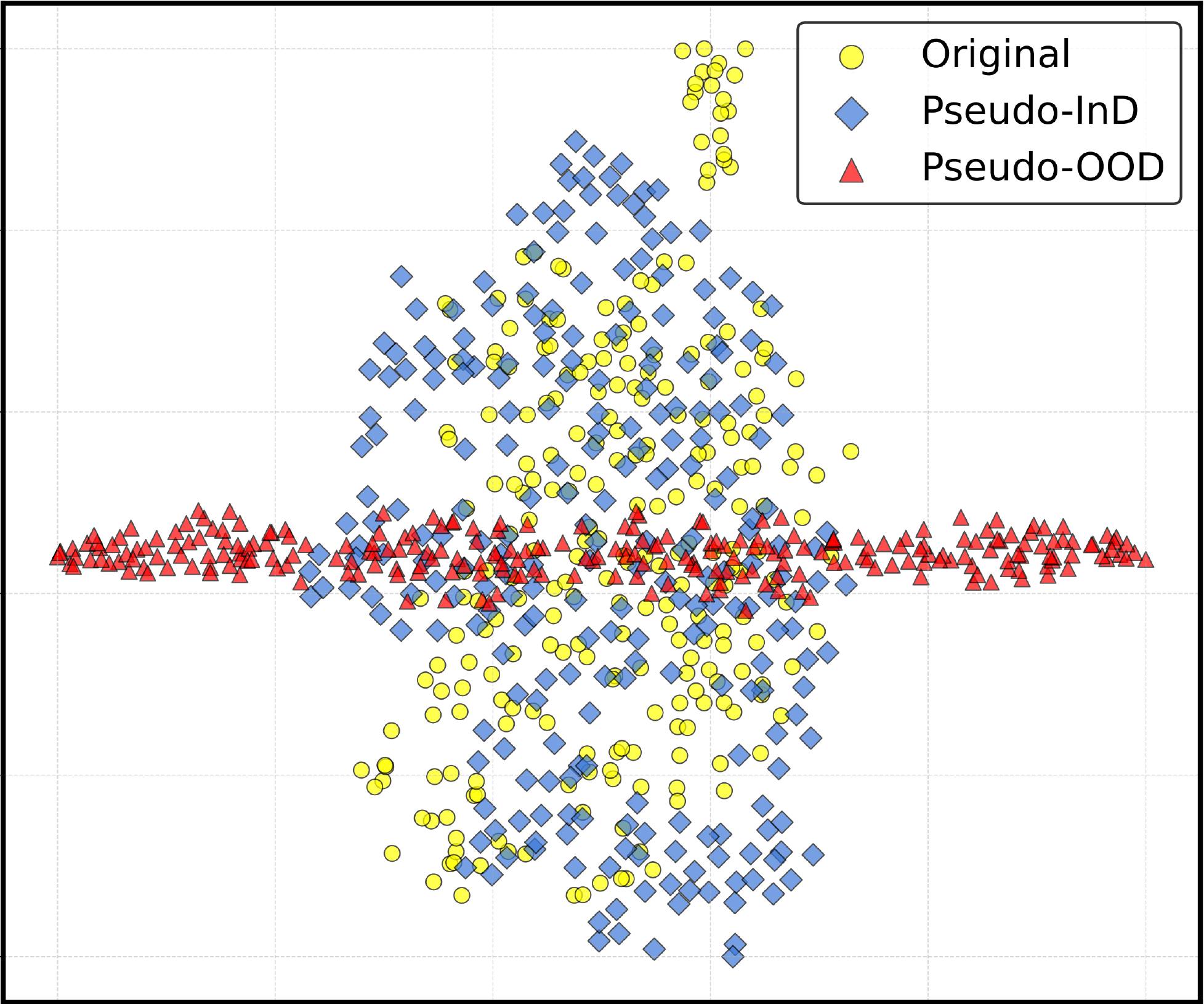}
        \centerline{(b) \footnotesize$\lambda$= 0.4}
    \end{minipage}
    \begin{minipage}[t]{0.32\linewidth}
        \centering
        \includegraphics[width=\textwidth]{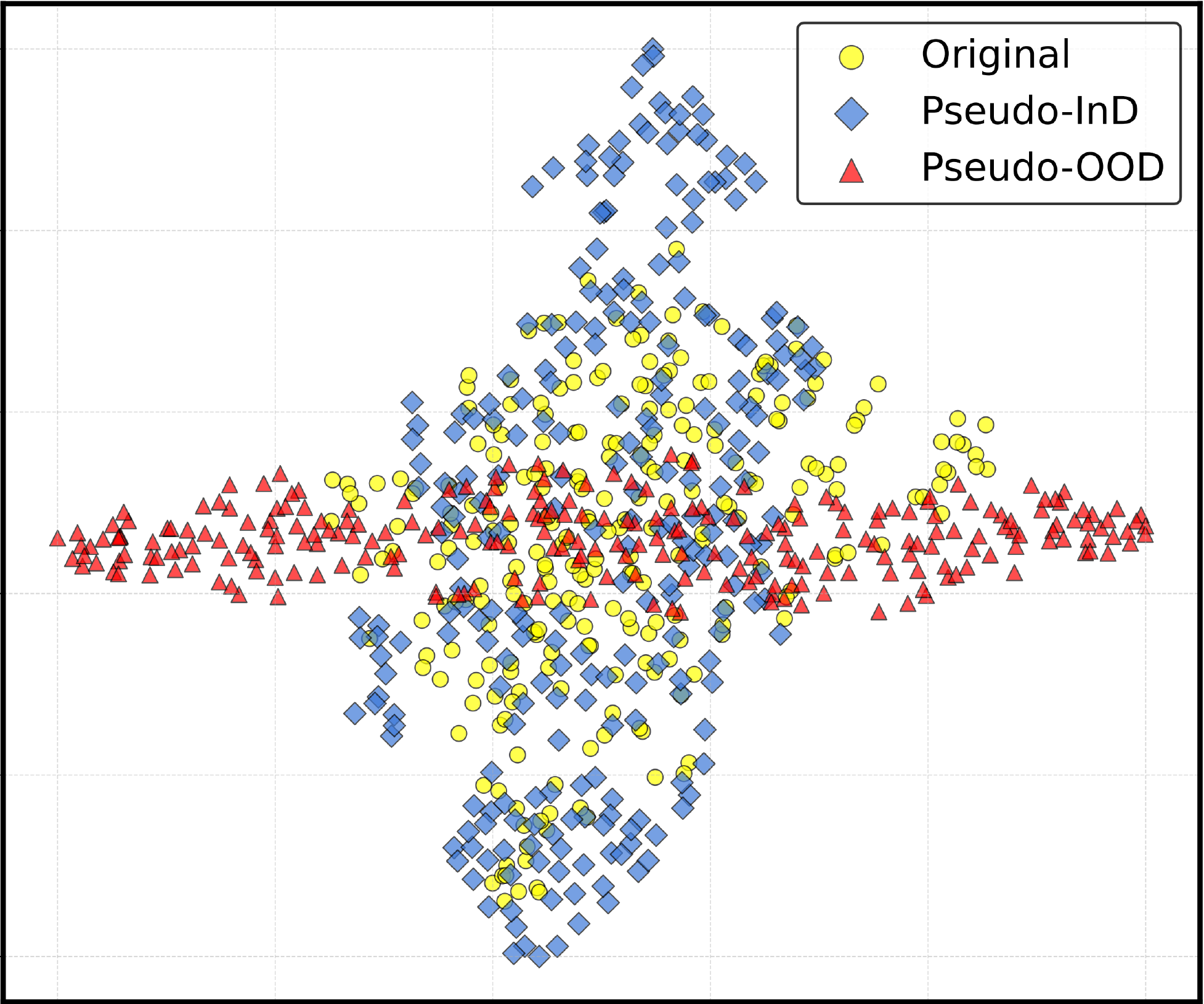}
        \centerline{(c) \footnotesize$\lambda$= 0.7 }
    \end{minipage}
    \caption{ t-SNE visualization of the original, pseudo-InD and pseudo-OOD representations. }
    \label{t-sne}
    \vspace{-5mm}
\end{figure}

The t-SNE visualization in Fig. \ref{t-sne} shows  the distribution of pseudo-InD representations predominantly extends vertically, while pseudo-OOD expands mainly horizontally. Moreover, as the values of \(\lambda\) increase, the pseudo representations deviate further from the original distribution. This observation indicates that our method is capable of effectively exploring the latent space. 
Unlike typical OOD generalization and OOD detection methods that account for only one type of distribution shift,
our \sysname{} takes the simultaneous occurrence of both shifts into account. As a result, our method achieves better performance to some extent. Additionally, open set domain generalization baseline SCONE performs poorly on \texttt{GOOD-CMNIST} and \texttt{GOOD-SST2} datasets but outperforms other baselines on another dataset. This suggests  this method has advantages on greater class imbalance and more significant shifts, but exhibits poorer robustness and adaptability. 

\begin{figure}
\centering
\includegraphics[width=1.0\linewidth]{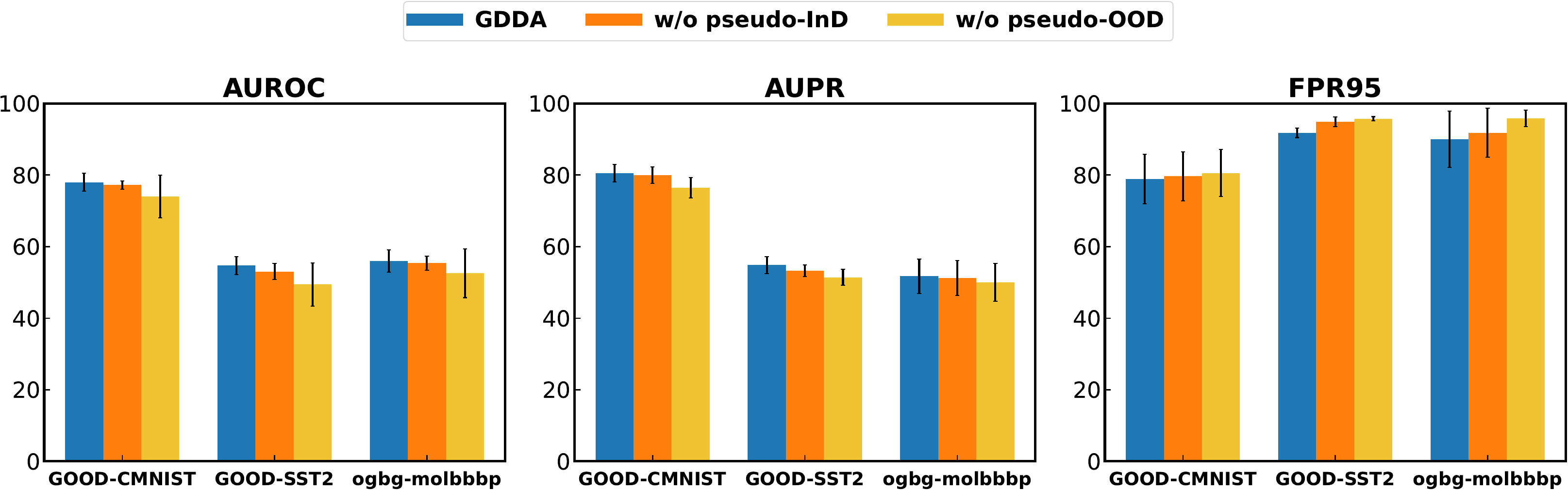}
\vspace{-7mm}
\caption{Ablation results on the three datasets.}
\label{fig:my_label}
\vspace{-5mm}
\end{figure}

\textbf{Ablation study.}
\label{sec:page}
We conduct ablation experiments to evaluate the contributions of  pseudo-InD and pseudo-OOD representations by removing each component individually. The conditions ``w/o pseudo-InD'' and ``w/o pseudo-OOD'' illustrate the impact of excluding pseudo-InD and pseudo-OOD. Fig. \ref{fig:my_label} shows that the AUROC score of ``w/o pseudo-InD'' exceeds that of ``w/o pseudo-OOD'', yet both scores are lower than \sysname{}. This suggests that pseudo-OOD  contributes more than pseudo-InD, and removing either module leads to performance degradation, highlighting the indispensability of both modules. 

\section{Conclusion}
\label{sec:prior}
To tackle the challenge of \textit{graph-level semantic OOD detection under covariate shift}, we propose two-phase \sysname{} framework. In the first phase, GDDA  disentangles graph representations into semantic and style factors. In the second phase, we generalize training factors to generate auxiliary pseudo-InD and pseudo-OOD representations with distribution-shift-controlled diffusion models to enhance the effectiveness of energy-based OOD detector. We demonstrate the superiority of our method compared with state-of-the-art baselines.